# NEMO: Extraction and normalization of organization names from PubMed affiliation strings

Siddhartha Jonnalagadda*, Philip Topham

Address: Lnx Research LLC, 750 The City Drive Suite 490, Orange, CA 92868, USA,

*Corresponding author at sjonnalagadda@lnxresearch.com

## Abstract

**Background.** We are witnessing an exponential increase in biomedical research citations in PubMed. However, translating biomedical discoveries into practical treatments is estimated to take around 17 years, according to the 2000 Yearbook of Medical Informatics, and much information is lost during this transition. Pharmaceutical companies spend huge sums to identify opinion leaders and centers of excellence. Conventional methods such as literature search, survey, observation, self-identification, expert opinion, and sociometry not only need much human effort, but are also non-comprehensive. Such huge delays and costs can be reduced by "connecting those who produce the knowledge with those who apply it". A humble step in this direction is large-scale discovery of persons and organizations involved in specific areas of research. This can be achieved by automatically extracting and disambiguating author names and affiliation strings retrieved through Medical Subject Heading (MeSH) terms and other keywords associated with articles in PubMed. In this study, we propose NEMO (Normalization Engine for Matching Organizations), a system for extracting organization names from the affiliation strings provided in PubMed abstracts, building a thesaurus (list of synonyms) of organization names, and subsequently normalizing them to a canonical organization name using the thesaurus.
**Results**: We used a parsing process that involves multi-layered rule matching with multiple dictionaries. The normalization process involves clustering based on weighted local sequence alignment metrics to address synonymy at word level, and local learning based on finding connected components to address synonymy. The graphical user interface and java client library of NEMO are available at http://lnxnemo.sourceforge.net.
**Conclusion:** NEMO associates each biomedical paper and its authors with a unique organization name and the geopolitical location of that organization. This system provides more accurate information about organizations than the raw affiliation strings provided in PubMed abstracts. It can be used for : a) bimodal social network analysis that evaluates the research relationships between individual researchers and their institutions; b) improving author name disambiguation; c) augmenting National Library of Medicine (NLM)'s Medical Articles Record System (MARS) system for correcting errors due to OCR on affiliation strings that are in small fonts; and d) improving PubMed citation indexing strategies (authority control) based on normalized organization name and country.

# 1. Background

In recent years, many systems have been developed for disambiguating author names from PubMed abstracts (1; 2). Torvik et al.(2) uses surface features such as title words, journal name, co-author names, medical subject headings, and language to estimate the probability that two PubMed articles that share the same author name were written by the same individual. Such an equivalent process for organization information is not available for PubMed abstracts.

Disambiguated author names are found to be directly useful for measuring research productivity of universities (3). However, such approaches have limitations (3) due to a) "extreme variance in the indication of affiliation information (full or abbreviated names in the original language or in English), or, conversely, identical indications for different institutions, for example universities located in the same city", i.e., **lack of highly precise systems for extraction of organization names**; and b) "indication of the affiliation to research centers which naturally fall within a larger institution (for example, laboratories, departments, institutes, medical centers, etc.)", i.e., **lack of systems that disambiguate organizations and map them to the naturally fitting largest organization or organization group**.

Considering that author disambiguation systems have achieved close to 100% accuracy (98% for Torvik et al.'s Author-ity system (2)), the missing piece for linking organizations to articles and their authors is a system that accurately extracts organization information from affiliation strings provided in PubMed, disambiguates them to canonical or most popular names, and maps the laboratories, departments, institutes, medical centers, etc. to a naturally fitting larger institution. *Disambiguating* and precisely *mapping* of institutions together will be referred to as normalization. Yu et al. (4) obtained detailed information about investigators by automatically extracting organizations and related entities from affiliation strings of the articles. Their system had an accuracy of 87% for extracting organization names. In addition, NextBio (5) and BioMedExperts (6) possess systems for extracting organization names from affiliation strings, but accuracy of these systems is not publicly known. Extracting organization names from free text is a well-studied problem (7). Recognizing organization names from the affiliation strings of PubMed abstracts is a different problem. Though the problem of normalization of organization names has been studied in open domains like Wikipedia and news articles as shown in (8), those systems had an accuracy of less than 80%. Thus, specialized tools are built for restricted domains such as gene mentions (9) and malignancy mentions (10). The aim of this research is to build a system that uses affiliation strings provided in PubMed to extract organization names with close to 100% accuracy and disambiguate the results.

### 1.2 Extracting organization names

**Classification:** There are mainly two relevant hierarchies of named entity types that have been proposed in the literature. The Bolt Beranek and Newman (BBN) categories (11), proposed in 2002 for a Question Answering task, consist of 29 types and 64 subtypes. Sekine's extended hierarchy (12), also proposed in 2002, consists of 200 subtypes. According to the BBN Hierarchy, our specific problem of extracting organization related named entities breaks down into identifying 4 major types of entities: a) Organization name – for classifying the names of the actual research groups, b) Geo-Political Entity name – names of country, state and city, c) Facility name – names of buildings and other man-made structures (Example: 1) Health Sciences West Bldg, 2) Room HSW1601), and d) Contact Information – Address, email and URL. While there has been much Named Entity Recognition (NER) work reported by researchers (for instances, in Mutual Understanding Conference - 7, Automatic Content Extraction - 2005 and Automatic Content Extraction - 2008), few (this work and (4-6)) have looked at the task of organization NER from the authors' affiliation strings.

**Approaches:** Early NER systems were mostly based on hand-crafted rules which in general have good performance, but they are labor intensive insofar as they require regular inputs from experts familiar with the text. Many NER systems for different entities have been replaced by supervised learning systems which need less effort from the developers; in such systems, rules are automatically created from the positive and negative examples. Unfortunately, such methods need large amounts of annotated data. To our knowledge, there are no publicly available annotated corpora for the different organization-related named entities discussed above. This and the need for a highly accurate system encouraged us to build a rule-based system.

**Data source:** A journal submitting papers to PubMed is required to send a single field representing the organization name (http://www.ncbi.nlm.nih.gov/entrez/query/static/spec.html). Usually this is the organization of the first author, and in some cases the organization which appears first in the journal abstract. In general, each article is associated with one organization. Currently, there is no standard style for listing authors' affiliation strings; it is a free form text field with some moderate cultural preferences to list them in the following order: institution, city, state, and country. However, there are wide variations in their style and format. For example, a person may list Department first and then Institution or vice-versa.

**Previous work:** In a previously known attempt at recognizing organization names from PubMed abstracts, Yu et al. (4) assumed that most of the affiliation strings in PubMed articles adhere to the following format: [address component], [address component], …, [country].[email], where an address component according to them could be an organization name, city, state or facility name. Yu et al. assumed that the affiliation strings have the name of the country explicitly and one of the address components contains one of the institution character n-grams to indicate that component to be an organization. However, there are many affiliation strings which do not obey these assumptions. In the affiliation string, "*Centaur Science Group, 1513 28th St NW, Washington, DC 20007 USA.* (PubMed ID: 16796054)", *Centaur Science Group* is an independent company that doesn't contain any of the key words of Yu et al. In addition, there are many organizations like *Mayo Clinic* which are present in multiple places like Rochester (Minnesota), Jacksonville (Florida), Scottsdale (Arizona), Phoenix (Arizona) and Fairfax (Virginia). For the purposes of deeper analysis, these organizations should be considered as different, which means that it is also necessary to extract the city and state names. Thus, the problem of NER for organization names and its attached features is non-trivial, but the output should have high precision and recall for being useful for large scale statistical analysis.

To create a system that recognizes organization names with high accuracy, we propose to apply rules at multiple levels, with each level gradually converting the unstructured input text into structured fields. This is discussed in section 2.1 and illustrated in Figure 1. Although we are dealing with citations written in or translated to English, about 10% of the institution names (such as *Universidad Central del Caribe* and *Carl von Ossietzky Universität Oldenburg*) remain in their native language.

### 1.3 Normalization of organization names

Until recently, there were quite a few algorithms for normalizing affiliation strings (13;14;15;16). Most of these algorithms are aimed at authority control to improve indexing of online bibliographic databases. Such applications have not been applied to PubMed which is the largest bibliographic database for biomedical domain. Although our main goal is to make such a system available in biomedical domain, our approach for normalization is also novel. In addition to dealing with variations because of spelling and abbreviation while resolving synonymy, we address synonymy through language translation and transliteration and we automatically identify basic hierarchical structure of organizations. We extend the edit-distance approach proposed by French et al. (13) by a) processing the affiliation strings at two levels – first to normalize variation caused by non-standard words and then the variation caused by non-standard combination of words, and b) using weighted local sequence alignment to disambiguate words and affiliation strings.

Summaries of our NER and normalization methods have been published (17;18); however, the present study adds details to make the algorithm reproducible. In addition to adding new methods and dictionaries, we have made the tool applicable to countries other than USA. Finally, we have released NEMO as a public web service/online application/graphical user interface which is available at http://lnxnemo.sourceforge.net/.

**Figure 1: Components of the NEMO system**

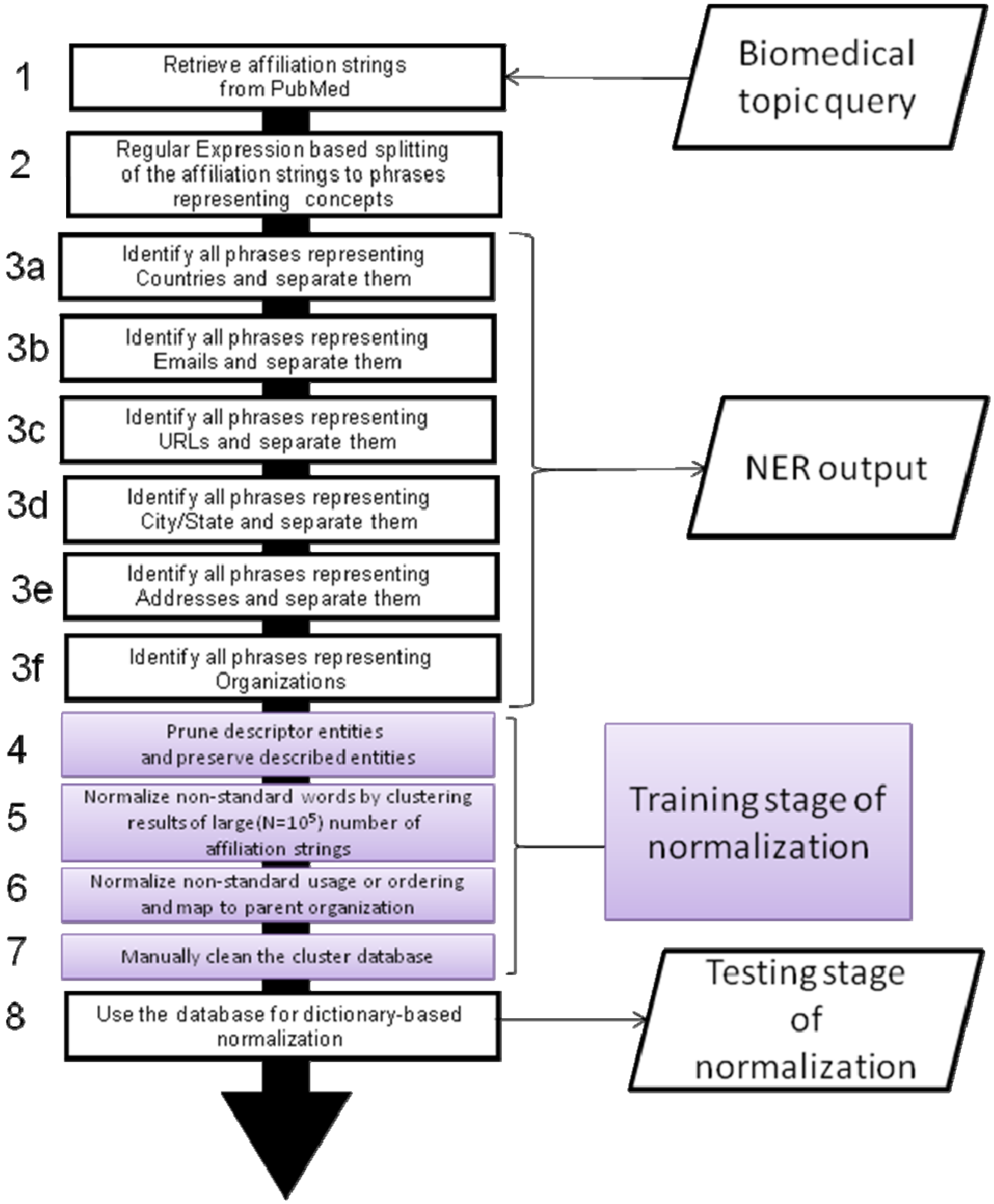

Step 1 constitutes the Information (affiliation strings) Retrieval stage where the affiliation strings are retrieved using NCBI eutils; steps 2 and 3 constitute the Named Entity Recognition stage; steps 4-6 constitute the training stage of Normalization; step 7 constitutes the post-processing stage of normalization (manual data cleaning); step 8 constitutes the testing (actual execution) stage of Normalization. Steps 4-7 are executed only once during the initial training stage where 103,557 random affiliation strings are used for unsupervised clustering to find canonical names for the organizations and map them to a parent organization name.

## 2. Methods

Figure 1 illustrates the architecture of NEMO. Starting from accepting a query for a biomedical topic, NEMO retrieves the affiliation strings using NCBI eutils (eutils.ncbi.nlm.nih.gov) [module 1], extracts the names of organizations and their locations [modules 2 and 3] and culminates in normalizing (uniquely identifying) it with an organization name, and associating it with a cluster of organizations that it collaborates with or is part of [modules 4-8].

### 2.1 Parsing affiliation strings to extract organization-related entities

The Vedic *Neti, Neti* golden rule [One must find the soul by analysis saying, "This is not it. This is not it."] holds true for extracting the organization names from the affiliation string. The other 3 types of named entities – geopolitical entity name, facility name and contact information – conform better to a general pattern than the most important entity, organization name. This is mainly because of the inherent volatility in the naming of an organization and its usage compared to that of a country, state, or city in which naming conventions are governed by social and political laws. Thus, we first find the phrases which represent only the country, email address, URL, state, city, and street address[1] (in that order) and then consider the left-over phrases to verify if they represent organization names. We store the geopolitical entity and other properties for future analysis. While this is the general framework, each sub task such as determining the name of the country requires use of multiple rules and multiple dictionaries. In total there are 30 different manually verified dictionaries or lists such as the dictionaries in Geoworldmap database (19), Internet domain – country mapping [IANA - http://www.iana.org/domains/root/db], stop words list (20), keywords for organizations (Table 1), keywords for addresses (Supplementary File 1) and zip code dictionary [http://www.zip-codes.com/zip-code-database.asp]. Table 2 details the steps in our process to extract organizations and related entities.

There are many more subtle rules and discussing all of them is beyond the scope of this paper. For example, the affiliation string: "*USDA, ARS, Aquatic Animal Health Research Unit, Auburn University, AL, USA*." may appear to be associated with *Auburn University*, but it is actually from *ARS* of *USDA*. The confusion is caused because *Auburn University* is the name assigned to a city in the USA with the zip code 36849. But there are many affiliation strings where Auburn University is an organization entity. In this case, when there is an ambiguous phrase that could be a city or an organization, we check for other phrases in the sentence that may represent a city. If this is true, the ambiguous phrase is considered as an organization name, otherwise it is treated as a city name.

[1] We don't distinguish between "facility name" [Examples: 1) Health Sciences West Bldg, 2) Room HSW1601 ] and "street address", and refer to both of them as address.

Our dictionaries were built initially for English and wherever applicable we translated them to additional languages using Google Translate (21). However, each dictionary item may have many synonyms in different languages and not all of them exist in the dictionary. If no information is extracted from a phrase in any of the steps in the extraction algorithm (Table 2), we translate individual phrases in the affiliation strings to English (using Google Translate) and execute all the steps (Figure 1, 3a-3f) again for that phrase. This approach is useful in cases where an address or organization keyword is missing from our dictionaries, but the English translation of the keyword exists within the dictionary list.

**Table 1: Keywords for organizations**

| | | | | |
|---|---|---|---|---|
| section | association | laboratorio | ortopedia | études |
| hospital | associates | laboratorio | ortopédico | biologie |
| hospitals | biology | estación | especialistas | projet |
| research | project | departamento | corporación | métabolisme |
| university | metabolism | departamentos | sustitución | programme |
| school | program | división | ingeniería | programmes |
| laboratory | programs | divisiones | desarrollo | unité |
| lab | unit | ingeniería | médicos | sportscare |
| station | bone | centro | instituciones | thérapie |
| department | sportscare | centro | compañía | groupe |
| departments | therapy | centros | empresas | centres |
| division | group | centros | servicio | orthopédie |
| divisions | centers | ciencias | clinique | orthopédiques |
| engineering | orthopedics | instituto | section | orthopédie |
| centre | orthopedic | institutos | hôpital | orthopédiques |
| center | orthopaedics | clínica | les hôpitaux | spécialistes |
| centers | orthopaedic | universidad | rechercher | Société |
| centres | specialists | medicina | université | de remplacement |
| sciences | corporation | sistema | école | génie |
| institute | replacement | dept. | laboratoire | développement |
| institutes | engineering | estudios | lab | médicaux |
| clinic | development | asociación | gare | institutions |
| college | medical | asociados | département | entreprise |
| medicine | institutions | biología | départements | entreprises |
| mc | company | proyecto | divisions | service |
| system | companies | metabolismo | génie | jôpitaux |
| dept. | service | programa | centre | abschnitt |
| studies | academy | programas | centres | krankenhaus |
| inc | pharma | unidad | sciences | krankenhäuser |
| inc. | université | hueso | institut | forschung |
| ltd | sección | SportsCare | instituts | universität |
| ltd. | hospital | terapia | clinique | schule |
| llc | hospitales | grupo | collège | labor |
| llc. | investigación | centros | médecine | bahnhof |
| llp | Universidad | ortopedia | système | abteilung |
| llp. | escuela | ortopédico | dept. | abteilungen |

**Table 2: Algorithm for extracting organizations and related entities**

<table>
<tr><td>Step 1: Data Extraction</td><td>1. Automatically query for the articles of interest using ESearch (45) and retrieve the corresponding PubMed identifiers<br>2. Fetch the related PubMed abstracts using EFetch (46) in XML-format<br>3. Parse the affiliation strings using the tag “affil str”</td></tr>
<tr><td>Step 2: Splitting the sentence into phrases and expanding the abbreviations</td><td>1. Find the phrases in the extracted affiliation data that match the following regular expression: (?<!(\\.))[^\\,\\;\\:\\@]+((\\,|\\;|\\:|$)(?=(\\p{Z}|[A-Z]|$))|(\\.)(?=(\\p{Z}|$)))<br>2. Detect abbreviations in the phrases found above, using the following regular expression: ((?<=(^|[^A-Za-z\\.]))(([A-Z]\\.?)([A-Z]\\.?)+)(?=([^A-Za-z\\.]|$))|([A-Z][A-Za-z]+\\.(\\p{Z}|$)))<br>3. Expand the abbreviations using abbreviations and their expansions found on the world wide web (47) and also store them locally for more than one use</td></tr>
<tr><td>Step 3: Country Detection</td><td>1. For each phrase found in step 2, detect if the entire phrase or a portion of it matches one or more variables associated with the country (such as country name, region name, city name, languages, email, country top-level domain, and zip codes)<br>2. If more than one country names are detected, assign the country detected from the phrase closest to the end<br>3. If the phrase only contains the country name or a variant of it, remove it from the list of phrases to be considered from the next step</td></tr>
<tr><td>Step 4: Email Detection</td><td>1. For each phrase remaining, detect if the entire phrase or a portion of it matches the regular expression: [a-zA-Z0-9!#$%&''*+,/=?^_`{|}~-])+(\\.[a-zA-Z0-9!#$%&''*+/=?^_`{|}~-]+)*@([A-Za-z0-9]([A-Za-z0-9-]*[A-Za-z0-9])?\\.)+[A-Za-z0-9]([A-Za-z0-9-]*[A-Za-z0-9])?<br>2. Store all the matched phrases or parts of the phrases as emails<br>3. If the phrase only contains the email, remove it from the list of phrases to be considered from the next step</td></tr>
<tr><td>Step 5: URL Detection</td><td>1. For each phrase remaining, detect if the entire phrase or a portion of it matches the regular expression: (?<=(\\s|^))(https?://|www\\.)[-\\w]+(\\.\\w[-\\w]*)<br>+(?![\\w\\.-]*@)<br>2. Store all the matched phrases or parts of the phrases as URLs<br>3. If the phrase only contains the URLs, remove it from the list of phrases to be</td></tr>
</table>

| | |
|---|---|
| | considered from the next step |
| Step 6: Address Detection | 1. For each phrase remaining, tag the phrase as an address depending on the identification whether or not: a) it contains a keyword for addresses, b) it contains numbers and directions, c) it contains keywords for organizations<br>2. If the phrase is an address, remove it from the list of phrases to be considered from the next step |
| Step 7: City and State Detection | 1. For each phrase remaining, determine the cities and states in by matching the entire phrase or a part of it by using the dictionary of: a) cities, b) states, and c) zip codes<br>2. If the phrase contains only a city or a state, remove it from the list of phrases to be considered from the next step |
| Step 8: Organization Detection | 1. For each phrase remaining, a) if the phrase following is a keyword for organizations or starts with "and", concatenate both of them, b) if the phrase contains a keyword for organizations and doesn't begin with numbers, tag it as organization, c) if there are no organizations added till now and it contains a place name or proper name, tag it as organization, d) if this is the last phrase and no organizations are found till now, tag it as organization. |

### 2.2 Normalization of organization names

**Described and Descriptor entities:** There are two distinct types of named entities related to an organization – those which uniquely identify with real world organizations (we refer to these as "described entities") and those which do not uniquely identify with real world organizations unless in the presence of a described entity. The primary role of the latter is to give more specific information about a described entity (thus, we refer to them as "descriptor entities"). All organizations containing a person name (recognized from an online dictionary), a place name (recognized from the dictionary of all major places in the world), or a directional modifier are recognized as "described entities". The rest of the organizations are "descriptor entities". There is at least one described entity per affiliation string. Examples of described entities are *Jerome Lipper Center for Multiple Myeloma*, *University of Texas* and *North-western University*. Examples of Descriptor entities include *School of informatics* and *Dept. of Biomedical Informatics*. Our glossary of person names is built using the following website: http://names.whitepages .com. This website is queried to learn whether a token is a person name. Both the positive and negative results are stored in our dictionary. After processing 103,557 random affiliation strings for extracting person names, the ambiguous names are manually removed. The process of updating the glossary continues perpetually while the software is being used. The glossary of places is built using the GeoWorldMap database (19).

**Resolving polysemy:** We addressed *polysemy* (same word having multiple senses) within the entity class of organization by mapping the extracted organizations to the least generic concept possible. For example, if an affiliation string is found to have *Mayo Clinic* as an organization and the only Geopolitical entity (GPE) recognized is USA (for the subtype country), we associated it with *Mayo Clinic* group of organizations in USA. However, if our NER process also recognized the city (say *Rochester)*, we associated it with the *Rochester* Branch of the *Mayo Clinic* group. The normalization process identified each described entity with a unique real world organization or a unique organization group as in the examples above. Since PubMed generally stores only the primary affiliation string of the first author, we associated each article with the normalized described entity that is estimated to have the highest number of articles in PubMed. For example, the article with PubMed ID – 15607955 has the following described entities after normalization: *Yale University School of Medicine*, and *Boyer Center for Molecular Medicine.* Out of the two, *Yale University School of Medicine* has higher number of articles; therefore, this article is associated with *Yale University School of Medicine*.

**Resolving synonymy:** One major challenge in normalizing organization names is to identify and replace Non Standard Words (NSWs). NSWs can be broadly classified as (22):

1. Miscellaneous – these are made of unconventional word and phrase boundaries, intentional informal spelling, URL and formatting abnormalities. The unconventional boundaries and URLs were dealt with in the NER phase. Informal spellings which might appear in conversational text do not appear in the organization names submitted to PubMed. Some affiliation strings often have formatting abnormalities such as the following: *From the *Division of Pediatric Cardiology and,* ***dagger****Department of Pediatrics, Ataturk University, Faculty of Medicine, Erzurum, Turkey.* (PubMed ID: 19262418, Date: 2009 Feb 28. [Epub ahead of print]), where the symbol † from the original affiliation string in the journal is replaced with the sequence “dagger” instead of getting deleted. While such abnormalities might arguably be useful for indicating the presence of multiple affiliations in the journal abstract, they prevent the abstracts containing them from being retrieved. These kind of mistakes are present even in recently manually indexed abstracts such as *Plant Polymer Research, USDA****,(dagger)*** *ARS, National Center for Agricultural Utilization Research, 1815 N. University St., Peoria, IL 61604, USA.* (PubMed ID: 19111748, Feb 2009). The NER stage does not remove these errors. These errors mainly appear at the stage of automatic scanning by the Medical Articles Record System (MARS) after escaping the notice of the Seek Affiliation program (23) and the manual supervisors of the National Library of Medicine’s indexing section.

2. Misspellings and abbreviations – Misspellings are dealt with at a later phase along with formatting abnormalities. Abbreviations need to be replaced by their full forms. For example, *University of California at Los Angeles Medical School* needs to replace *UCLA Medical School*. This problem is automatically solved (at the NER step – Table 2, step 2) by making sure that all the acronyms are expanded to their full form before tagging a phrase as an organization. For example, the organizations identified for the affiliation string “*Baylor College of Medicine,* ***USDA/ARS*** *Children's Nutrition Research Center, Department of Pediatrics-Nutrition, Houston, Texas, USA*. (PubMed ID: 12401707)” was expanded to *Baylor College of Medicine,* ***United States Department of Agriculture/Agriculture Research Service*** *Children's Nutrition Research Center* and *Department of Pediatrics-Nutrition*.

Thus, spelling variations and formatting mistakes can cause NSWs to appear in organization names that are extracted automatically. These two along with the lack of consensus in the choice of words when referring to the organizations are responsible for synonymy (different words for the same concept) in organization named entities. Table 3 shows an example of synonymy. The task of Normalization is to map all these 10 organization mentions to the same concept – Washington University School of Medicine. One common approach to solving this synonymy is to compare the recognized organization name against a list or dictionary of organization names. This approach is used in gene normalization systems (24) where the systems find the Entrez Database Gene identifiers for human genes or gene products mentioned in PubMed abstracts. Unlike gene names, organization names are volatile. Many organizations get renamed and some organizations become non-operational every few years. We did not notice a public database of organizations that is also maintained. In the present study, we propose a mechanism to automatically build a database of organization clusters, OrgDB, from 103,557 randomly selected affiliation strings from PubMed published between the years 1998 and 2008. Then, we used automatic techniques to match the database entries in order to normalize the organization names.

**Table 3: Example of Synonymy**

| PMID | |
|---|---|
| 20740553 | Washington University School of Medicine |
| 20653866 | Washington University School of Medicine and St. Louis Children's Hospital |
| 20720053 | School of Medicine |
| 20506172 | Washington University School of Medicine at Barnes-Jewish Hospital |
| 18762643 | Barnes-Jewish Hospital at Washington University School of Medicine |
| 20720053 | Washington University |
| 17182886 | Washington University School of Medicien* |
| 12056922 | Washington University School of Medicine and Metropolitan St. Louis Psychiatric Center |
| 11195749 | Barnes Retina Institute and Washington University School of Meidcine* |
| 10784594 | Division of Gastroenterology Washington University School of Medicine St. Louis |

* ibid

Two other important causes of synonymy in the non-English names of organizations are a) they might be referred sometimes by their English translation (Example: University of Chile in PMID: 20737243) and other times by the original name (Example: Universidad de Chile in PMID: 20735270); and b) the diacritic marks might be preserved (Example: Universitat **Autònoma** de Barcelona in PMID: 20735775) or removed (Example: Universitat **Autonoma** de Barcelona in PMID: 20734978). To disambiguate this synonymy, we translated each non-English organization name to English using Google Translate (21) and removed all diacritic marks.

After the NER step and translation and transliteration to English, we use clustering to automatically build a dictionary for organizations and their synonyms [modules 4-7], and then dictionary based matching [module 8] using the dictionary we built in the previous step.. We store the thesaurus of organizations as a database, OrgDB, starting from the organization names parsed from the affiliated sentences. Each entry in the OrgDB database is a cluster that has the following features: a) a centroid string, b) a list of all organizations in the cluster, c) a matrix containing inter-component distance using the string similarity metric, d) the PubMed IDs of the articles containing at least one organization name from the cluster, e) the city, state and country of the cluster.

**Clustering**: Our approach to clustering is agglomerative and partitional. OrgDB is initially empty and each new organization name along with the related information is added to OrgDB and to one of the clusters already present in the database if the organization name is sufficiently close to its centroid (according to a threshold of edit distance as defined below). After adding a new organization to a cluster, the centroid is recomputed. Usually centroids are calculated by taking the average of the vector representations of the elements in the cluster. In a Euclidean vector space, it is an easy proof using Fermat's theorem on stationary points that the centroid is the point that has the smallest sum of the squares of the Euclidean distances from each of the points in the cluster. To represent organization names in a continuous vector space similar to the document-term vector space used in information retrieval and also encode the order of the terms in the name of an organizations, one may need to use an $O(N^L)$ dimensional space, where N is the size of the vocabulary used in organization names (which is of the order 1000) and L is the maximum number of words in an organization string being considered(which is of the order 10). Since a vector space of dimensionality in the order of $10^{30}$ is prohibitively huge, we are using edit distance between strings as our kernel function for calculating distance between two organization names without representing the organization names in a vector space. Since we do not have a numerical representation for the organization names, an organization name from the cluster which is the best approximation for an ideal centroid is chosen as centroid.

The centroid is chosen to be the organization whose name has the least sum of edit-distances from the names of all organizations in the cluster. The GPE of the cluster is the set union of the GPE of all organizations; i.e., if the parsing process is not able to identify the city of one of the organizations while being able to identify the city in another organization within the same cluster, then the city of the latter organization becomes the city associated with the cluster.

Each affiliation string is processed through the NER mechanism in order to get all the organization names along with their GPE. Each described entity among those organizations is compared with the centroid strings of the clusters from OrgDB which have the same GPE. The distance metric that is used in this case is discussed below. If one of the subtypes of the GPE – city, state, or country is missing, then all the clusters in OrgDB with the same set of remaining subtypes are used for edit-distance comparison. If the distance metric suggests that an organization is sufficiently close to the centroid of a cluster, then the organization is added to the cluster. If no cluster is close enough to the organization, a new cluster is added to OrgDB having the organization as the only component. Thus, we obtained clusters of organizations with each cluster having components with minor variations at a lexical level. This clustering step is performed for a sufficiently large number of affiliation strings (N=103,557) so that most of the organizations and their statistical distribution are known in a reasonable amount of time.

**String Similarity:** For comparing the variants in the organization names, we are inspired by the biological sequence alignment algorithms that have been used recently for text mining applications such as sentence paraphrasing (25;26). There are two categories of sequence alignment: Global sequence alignment as implemented by the Needleman-Wunsch (NW) algorithm (27), and Local sequence alignment as implemented by the Smith-Waterman (SW) algorithm (28). Needleman-Wunsch algorithm is a dynamic programming algorithm for pair-wise global sequence alignment, i.e., it is used to find the best possible alignment between two sequences. The Smith-Waterman algorithm is similar to the Needleman-Wunsch algorithm except that it seeks optimal sub-alignments instead of a global alignment and, as described in the literature, it is well tailored for pairs with considerable differences in length and type. Table 4 demonstrates the use of these algorithms; the first string is the string that we are interested in normalizing while the second string is from OrgDB. We are calculating the NW and SW scores as implemented using Neobio software (29). We use the basic scoring scheme of an award of 1 for match and a penalty of 1 for both mismatch and gap. The second example demonstrates that, (also shown by Corderio et al. (26) in the context of paraphrasing sentences), global sequence alignment may not be suitable for the purpose of comparing organization names since it can classify related strings as different. Although local sequence alignment may seem to suit our purpose, there are cases where it can classify unrelated strings to be similar as in Table 4, c&d. The local sequence alignment wrongly identified WOMEN AND CHILDREN HOSPITAL LOS ANGELES with CHILDREN HOSPITAL LOS ANGELES instead of WOMEN'S & CHILDREN'S HOSPITAL. This is why we need a different mechanism of comparison that initially has a strict scheme for comparing organization names (i.e, it holds off on the matching process until enough information is available to make a concise and reliable match call). This kind of approach can be aptly called "recalculation through self-training" and is recently adopted in building an efficient natural language parser (30) which is currently one of the best parsers available in biomedical domain. Such a method of using local information from the training data to further enhance the value of the training set is referred to as "local learning" in the field of Artificial Intelligence (AI) (31).

**Table 4: NW and SW scores**

**a: pair 1**

| | |
|---|---|
| **First String** | ST. LUKE'S AND ROOSEVELT HOSPITAL |
| **Second String** | ST LUKES ROOSEVELT HOSPITAL |
| **NW match** | ST. LUKE'S AND ROOSEVELT HOSPITAL<br>ST- LUKE-S —— ROOSEVELT HOSPITAL |
| **NW score** | 21 |
| **SW match** | ST. LUKE'S AND ROOSEVELT HOSPITAL<br>ST- LUKE-S —— ROOSEVELT HOSPITAL |
| **SW score** | 21 |

**b: pair 2**

| | |
|---|---|
| **First String** | CHARLOTTE ORTHOPEDIC HIP AND KNEE CENTER AND CHARLOTTE ORTHOPEDIC RESEARCH INSTITUTE |
| **Second String** | CHARLOTTE'S ORTHOPEDIC HIP AND KNEE CENTER |
| **NW match** | CHARLOTTE-- ORTHOPEDIC HIP AND KNEE CENTER AND<br>CHARLOTTE'S ORTHOPEDIC HIP AND KNEE C------E-----N--<br>CHAROLETTE ORTHOPEDIC RESEARCH INSTITUTE<br>-------------------------T--------------------E--R-------------------- |
| **NW score** | -6 |
| **SW match** | CHARLOTTE-- ORTHOPEDIC HIP AND KNEE CENTER<br>CHARLOTTE'S ORTHOPEDIC HIP AND KNEE CENTER |
| **SW score** | 38 |

Global sequence alignment may not be suitable for our purpose of comparing Organization mentions since it can classify related strings as different

**c: pair 3**

| | |
|---|---|
| **First String** | WOMEN AND CHILDREN HOSPITAL LOS ANGELES |
| **Second String** | CHILDREN HOSPITAL LOS ANGELES |
| **SW match** | CHILDREN HOSPITAL LOS ANGELES<br>CHILDREN HOSPITAL LOS ANGELES |
| **SW score** | 29 |

**d: pair 4**

| | |
|---|---|
| **First String** | WOMEN AND CHILDREN HOSPITAL LOS ANGELES |
| **Second String** | WOMEN’S & CHILDREN'S HOSPITAL |
| **SW match** | WOMEN AND CHILDREN-- HOSPITAL<br>WOMEN’ S & CHILDREN'S HOSPITAL |
| **SW score** | 17 |

The local sequence alignment wrongly identified WOMEN AND CHILDREN HOSPITAL LOS ANGELES with CHILDREN HOSPITAL LOS ANGELES instead of WOMEN’S & CHILDREN'S HOSPITAL.

**Tight String Similarity (TSS):** We are using the Levenshtein distance (32) (the most commonly used edit distance metric) between the two organization names, not at the character level but at the word level. We remove all the words that are defined as stop words in our dictionary (20), as the presence or absence of a stop word does not change the identity of an organization. Two given words of non-zero length are considered same if they score more than 0.85 on the word similarity score (WS, defined in equation 1). The threshold of WS was chosen optimally to be 0.85 in order to prevent mismatching of more than1 letter for every 7 letters (1-1/7~0.85)

$$WS(a, b) = \frac{\text{Smith-Waterman Alignment score of a and b (34)}}{\text{Average length of a and b}} \quad (1)$$

For the Levenshtein calculation, the penalty for a gap of a word is the length of the word and the penalty for a mismatch between two words is the sum of their lengths. This penalty was chosen because larger words in general have more information; hence the penalty should be proportional to the length of the words. Since the current step is “Tight” String Similarity, we need to have a similarity match that is tight (i.e, strict) enough to prevent classifying different organizations as similar. Thus, we chose a conservative threshold of 4 (i.e., two organization phrases are similar if their Levenshtein scores are not more than 4). Using this similarity metric, we associate the described organizations identified from the NER process to one of the clusters in OrgDB if such a cluster exists, otherwise a new cluster in OrgDB is formed.

**Recalculation**: Because TSS assures standardization only on the words and not on the whole sentence, it mainly addresses the synonymy caused by NSWs. The organizations represented by two or more different clusters might still represent the same organization because of the lack of consensus in the choice of words while referring to the organization. Example: “*The David Geffen School of Medicine at the University of California”* and “*DG School of Medicine at the University of California at Los Angeles”*. So, in the recalculation step we find all the organizations related to the centroid of the present cluster. The algorithm is based on finding the connected component containing a vertex (33).

OrgDB is equivalent to an undirected graph, OrgG, with the vertices as the different clusters. An edge exists between two vertices (clusters) only if they are not from different cities or states and their corresponding centroids a and b score more than 0.90 on the Extended Smith-Waterman Score (ESS, defined in equation 2). This means that two organization names are related, if roughly for every 10 letters in the smaller name, only 1 letter can mismatch. This is still a conservative threshold, thus minimizing false positives or type-1 error.

$$\text{ESS}(a, b) = \frac{\text{Smith-Waterman Alignment score of a and b (34)}}{\text{Minimum of the lengths of a \& b}} \qquad (2)$$

If one of the two strings contains most of the other string (e.g, *David Gaffen School of Medicine* and *DG School of Medicine)*, then their ESS would be more than 0.90 and there will be an edge between the vertices corresponding to the clusters of these two strings in OrgG. Our recalculation step is equivalent to finding the “connected component” that contains the vertex corresponding to the cluster that contains the organization being normalized. In this paper, we will use the phrase “connected component” to refer to “connected component containing the vertex we are currently interested in”. The connected component is calculated by the breadth-first approach as elaborated in Table 5.

**Table 5: Algorithm for Recalculation**

| |
|---|
| 1. The connected component initially just contains the vertex (cluster) we are concerned with |
| 2. Iteratively visit each unvisited vertex (cluster) at depth 1 from the root (the initial vertex or cluster) |
| 3. Add all the vertices (clusters) adjacent to it and are not already in the connected component |
| 4. After the connected component has finished adding all the vertices (clusters) till the depth of 2 from the root, we consider only those vertices (clusters) which have an organization that is mentioned in the same PubMed article along with another organization in one of the vertices (clusters) already in the connected component. Such a pruning was observed by us to prevent adding wrong vertices (clusters) to the connected component |

**Cleaning process:** OrgDB is manually cleaned by: 1) removing all the clusters where the centroid is a descriptor entity, 2) merging of two clusters if the output of normalization of the centroids refers to the same organization, 3) expanding unambiguous abbreviations, 4) correcting the spelling mistakes in the names of centroid of clusters, and 5) repeating the Recalculation step.

**Running/testing stage:** An example of the dictionary matching process [Figure 1, module 8] is exemplified in Figure 2 and Table 6. The input affiliation string is "*Duke University Medical Center and Duke Clinical Research Institute, Durham, NC 27710, USA.*". We get the organizations O1-O20 through the process of first clustering and then finding the connected components till a depth of 2 from the root vertex O1. To demonstrate the need for the pruning step, we consider expanding O19 which is a leaf. The organizations adjacent to it are: *Durham Veterans Affairs Medical Center* and *Veterans Affairs Medical Center*. An examination of the PubMed IDs associated with articles in the clusters that are in the connected component of the above two organization vertices revealed that they did not collaborate (in any of the 103,557 publications we analyzed) with any organization in the connected component. Thus, we did not add these organizations to the connected component. Step 4 is continued for the rest of the organizations. The connected component gives the set of all the synonyms of the organization to be normalized. This set (Table 6) is further sorted in the decreasing order of number of components in the corresponding cluster in OrgDB. Depending on the objectives of normalization, the criterion to choose the representative organization varies. Currently, we chose the centroid string of the cluster with the largest number of publications to be the normalized name so that the normalized organization name is the name that is used most frequently. According to this criterion - *Duke University Medical Center* becomes the normalized name for *Duke University Medical Center and Duke Clinical Research Institute.*

**Table 6: Organizations in Connected Component for the example in Figure 2**

| Symbol | Organization Name |
|---|---|
| O1 | Duke University Medical Center and Duke Clinical Research Institute |
| O2 | Duke Clinical Research Institute |
| O3 | Duke University Medicalcenter |
| O4 | Duke University Medical Center |
| O5 | Duke University |
| O6 | Department of Biostatistics and Bioinformatics and Duke Clinical Research Institute |
| O7 | Duke University Medical Center Durham |
| O8 | Duke University Medical Center Duke University |
| O9 | Department of Psychiatry and Behavioral Sciences Duke University Medical Center Durham |
| O10 | Box 3709 Duke University Medical Center |
| O11 | Division of Gastroenterology Duke University Medical Center Durham |
| O12 | Veterans Affairs and Duke University Medical Centers |
| O13 | Duke University Eye Center |
| O14 | Duke University Health System |
| O15 | Duke University School of Nursing |
| O16 | Duke University School of Medicine |
| O17 | Duke University Hospital |
| O18 | Duke University Pain Prevention Center |
| O19 | Duke University and Durham Veterans Affairs Medical Center |
| O20 | Preston Robert Tisch Brain Tumor Center at Duke University |

**Figure 2: An Example. Input:** Duke University Medical Center and Duke Clinical Research Institute, Durham, NC 27710, USA. (PMID: 16849888 )

**a** *NER*

- *Organization*: Duke University Medical Center and Duke Clinical Research Institute (O1)
- *Country*: USA
- *State*: North Carolina
- *Zip Code*: 27710
- *City*: DURHAM

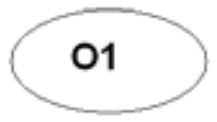

*Adding to OrgDB or OrgG*

- Add new cluster for O1 (*Recalculation Step 1*)

**b** **Recalculation Step 1**

- Organizations adjacent to O1 are Duke Clinical Research Institute(O2), Duke University Medicalcenter(O3), Duke University Medical Center(O4), and Duke University(O5)
- Add these to the connected component

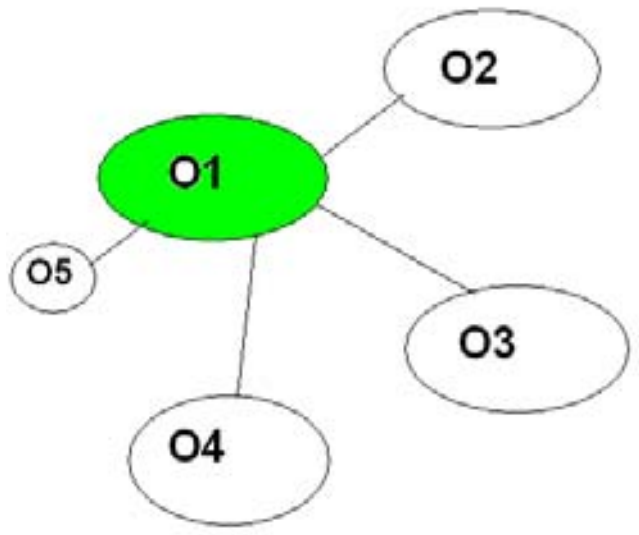

**c** Recalculation Step 2

Expand all the nodes at level 2. For example, O2. An organization adjacent to O2 is Department of Biostatistics and Bioinformatics and Duke Clinical Research Institute (O6).

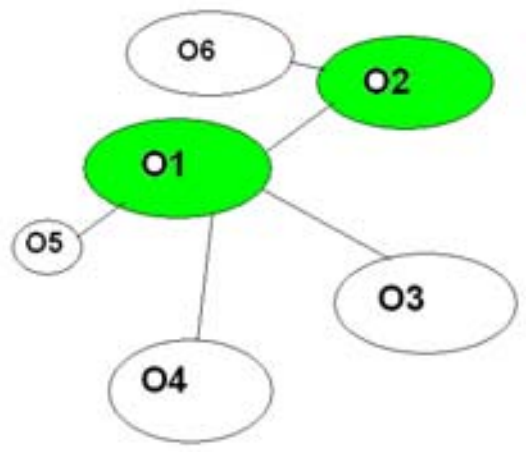

**d** **Recalculation Step 3**

- Do the same thing for O3, O4 and O5. We get 14 more organizations in the connected component – O7 through O20 listed in Table 6.

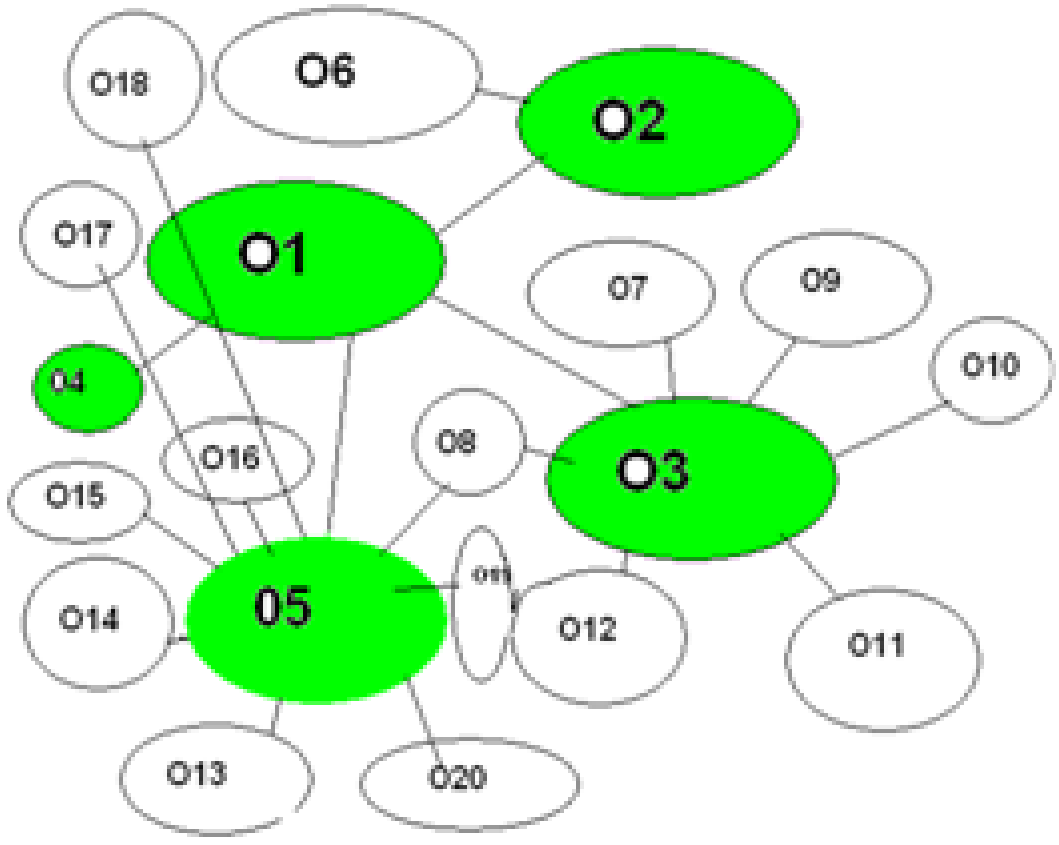

## 3. Results

As discussed above, we initially built the system for articles associated with USA organizations only. We then systematically extended our system to handle articles from more than 100 countries. A list of these countries is provided in Table 7. If an affiliation string is from USA, the component of NEMO built for USA is used; otherwise the component built for other countries is used. Thus, we have separate evaluations for: a1) determining whether the country is USA or not and extracting organizations and related entities from USA affiliation strings (stage 1); a2) normalization of organizations extracted from USA affiliation strings (stage 2); b1) determining the name of the country from the affiliation string (stage 3); b2) extracting organizations and related entities from all affiliation strings (stage 4); and b3) normalization of organizations extracted from all affiliation strings (stage 5). Once the corresponding component of NEMO is implemented, an evaluation is performed. Thus, rather than building the system all at once, the lifecycle of NEMO is divided into 5 separate stages. Each stage ends with an evaluation phase which is equivalent to the testing phase in software development life cycle. To make sure that our results are not high only because of over-fitting, we used different sets of affiliation strings for each of the different evaluations. Table 8 presents the results for different stages of the evaluation process assessed using precision, recall and f-score measures. Overall, the results (precision, recall and f-score) are excellent. Table 9 gives examples of false positives and false negatives for the NER stage. We did not quantitatively measure the recall of normalization. NEMO was often capable of normalizing organization names better than our human annotators as demonstrated in Table 10 for USA organizations and in Table 11 for organizations for non-USA countries.

**Table 7: Countries handled by NEMO**

| | | |
|---|---|---|
| Albania | Germany | Pakistan |
| Algeria | Greece | Panama |
| Argentina | Guatemala | Peru |
| Armenia | Honduras | Philippines |
| Australia | Hungary | Poland |
| Austria | Iceland | Portugal |
| Azerbaijan | India | Qatar |
| Bahrain | Indonesia | Romania |
| Bangladesh | Iran | Russia |
| Barbados | Ireland | Saudi Arabia |
| Belarus | Israel | Serbia |
| Belgium | Italy | Singapore |
| Bosnia and Herzegovina | Jamaica | Slovakia |
| Brazil | Japan | Slovenia |
| Bulgaria | Jordan | South Africa |
| Cambodia | Kazakhstan | South Korea |
| Canada | Kenya | Spain |
| Chile | Kuwait | Sri Lanka |
| China | Latvia | Sweden |
| Colombia | Lebanon | Switzerland |
| Costa Rica | Lithuania | Taiwan |
| Croatia | Malaysia | Thailand |
| Cuba | Mali | The Former Yugoslav Republic of Macedonia |
| Cyprus | Mexico | Trinidad and Tobago |
| Czech Republic | Moldova | Tunisia |
| Denmark | Morocco | Turkey |
| Dominican Republic | Myanmar | Ukraine |
| Ecuador | Netherlands | United Arab Emirates |
| Egypt | New Zealand | United Kingdom |
| El Salvador | Nicaragua | United States |
| Estonia | Nigeria | Uruguay |
| Finland | North Korea | Uzbekistan |
| France | Norway | Venezuela |
| Georgia | Oman | Vietnam |

**Table 8: Results of NEMO for different stages**

| Disease Area | Task | Precision | Recall | F-score |
|---|---|---|---|---|
| 6042 affiliation strings related to “Atrial Fibrillation” | Detect if the country is USA | 100.0% | 99.9% | 100.0% |
| 6042 affiliation strings related to “Atrial Fibrillation” | NER for organization names from USA | 99.2% | 99.9% | 99.6% |
| 4135 affiliation strings related to “Antiangiogenesis” | Normalization of organizations from USA | 99.5% | -- | -- |
| 4910 affiliations strings related to “Glaucoma” | Detect the country[1] | 100.0% | 99.5% | 99.8% |
| 4000 affiliations strings related | NER for organizations from | 100.0% | 97.5% | 98.7% |
| to "Staphylococcus Aureus" | all countries[2] | | | |
| 1000 affiliation strings related to "Diabetes" | Normalization of organizations from all countries[3] | 97.9% | -- | -- |

[1] The other known system by Yu et al. (4) has a precision of 94.0%, recall of 92.1% and f-score of 93.0% for detecting country names on a different dataset. The slight decrease in recall is because we are detecting only the countries listed in Table 7

[2] The other known system by Yu et al. (4) has a precision of 86.8%, recall of 91.3% and f-score of 89.0% on a different dataset for all countries. NEMO also achieved f-scores of 93.7% and 94.9% in identifying the names of states and cities respectively. The lower performance in these categories can be attributed to inherent ambiguity between these entity types.

[3] A few interesting associations similar to those in Table 10, but at international level are presented in Table 11. Verifying these associations needed domain expertise or internet search.

**Table 9: Extraction of organization names from USA - Examples**

| PubMed ID | Sentence | Organizations Recognized | Precision% | Recall % |
|---|---|---|---|---|
| 19121441 | MetroHealth Campus, Case Western Reserve University, Cleveland, Ohio, USA. *XXX* @metrohealth.org | 1. Case Western Reserve University | 100 | 50 |
| 16475045 | VA Boston Healthcare System and Beth Israel Deaconess Medical Center, Harvard Medical School, Boston, MA, USA. *XXX* @bidmc.harvard.edu | 1. VA Boston Healthcare System and Beth Israel Deaconess Medical Center<br><br>2. Harvard Medical School | 100 | 100 |
| 19017662 | Department of Medicine, Cardiology Division, LSU Health Sciences Center, Shreveport Louisiana, USA. | 1. Department of Medicine<br>2. Cardiology Division<br>3. LSU Health Sciences Center<br>4. Shreveport Louisiana | 75 | 100 |

**Table 10: Interesting mappings of organization names from USA**

| Organization from NER | Output of Normalization |
|---|---|
| National Cancer Institute | National Institutes of Health |
| Barbara Ann Karmanos Cancer Institute | Wayne State University |
| H. Lee Moffitt Cancer Center and Research Institute | University of South Florida |
| The Methodist Hospital, Texas | Baylor College of Medicine, Texas |
| Case Comprehensive Cancer Center | Case Western Reserve University |
| University Of North Carolina School of Pharmacy | University Of North Carolina At Chapel Hill |
| Wills Eye Hospital | Thomas Jefferson University |
| Vanderbilt-Ingram Cancer Center | Vanderbilt University School Of Medicine |
| Keck School of Medicine | University of Southern California |
| Brown Medical School | Rhode Island Hospital |
| Eyetech Pharmaceuticals, Inc | Imclone Systems Incorporated |

**Table 11: Interesting mappings of organization names from all countries**

| Organization from NER | Output of Normalization |
|---|---|
| University Hospital | Lund University |
| Rosalind Franklin University of Medicine And Science | The Chicago Medical School |
| Academic Medical Center | University of Amsterdam |
| Kuopio University Hospital | University of Kuopio |
| Hadassah Hebrew University Medical Center | The Hebrew University of Jerusalem |
| David Geffen School of Medicine | University of California, Los Angeles |
| Irvine Medical Center | University of California, Irvine |
| Charite-University Medicine | University Medicine Berlin |
| John A. Burns School of Medicine | University of Hawaii |
| Chung-Ho Memorial Hospital | Kaohsiung Medical University |
| Bispebjerg University Hospital | Copenhagen University Hospital |
| The National University Hospital | Rigshospitalet |
| University Hospital | Uppsala University |
| Feinberg School of Medicine | Northwestern University Medical School |

All the above discoveries were solely based on the organization information in OrgDB. Because we are using the sophisticated “connected components based recalculation” as opposed to the straightforward string similarity, we enjoy the advantage of discovering a richer set of synonyms than naïve approaches. For example, *Harvard Medical School* appeared in the synonym sets of most of its affiliates listed in http://hms.harvard.edu/admissions/default.asp?page=affiliates. Since *Harvard Medical School* has the highest rank in OrgDB, according to our chosen criterion of Normalization, all these organizations got automatically identified with it.

## 4. Discussion

**Evaluation methods:** Each of our evaluations involved at least two expert data analysts (employees of Lnx Research – http://www.lnxpharma.com). Since the evaluation was done at 5 different time intervals over the last one year, we did not necessarily have the same set of evaluators for each stage of evaluation. Each analyst judged the system's output on each category as either true positive, false positive, false negative, or true negative. When there was a disagreement in a judgment, the analysts discussed it until all of them finally agreed. We believe that biasing the evaluators with the output of the system was reasonable because even in cases where there are multiple correct outputs, it is acceptable for the system to generate one of the acceptable outputs. Creating a gold standard for the sake of evaluation would be much more expensive, especially in cases when there can be more than one correct output. We calculated the precision, recall and f-score values (34) using the true positive, false positive and false negative values for each category obtained after consensus of the analysts.

**Clustering for disambiguation:** Since the centroid is the point that has the least sum of the squares of the distances from each vertex, we choose the organization that has the least sum of edit distances with each organization name in the cluster as the approximation for centroid. While this training process has a time complexity of O(k*N) where k is the number of clusters (same time complexity as k-means, a standard partitional clustering algorithm), identifying whether a new affiliation string belongs to one of these k clusters (dictionary matching phase [module 8 in Figure 1]) has a time complexity of only O(k). Compared to k-means, our algorithm has the major advantage that k needn't be known apriori. We can obtain the same clusters using agglomerative hierarchical clustering (time complexity = $O(N^2) >> O(k*N)$) and splicing the dendrogram at a depth chosen to maintain the threshold of edit distance we fixed. In our approach, the value of the threshold of the edit distance needs to be known apriori.

**Limitations:** The limitation of the organization name extraction component is the expansion of abbreviations using a glossary. The two-character and three-character abbreviations are usually very ambiguous. For example, in "Department of Experimental Therapy of ARS" 'ARS' was initially being replaced with 'Agricultural Research Service'. We currently deal with ambiguous acronyms by not expanding them. When we realized that the ARS for this phrase refers to an organization in China, we deleted ARS in the glossary. In the next version of NEMO, we wound expand each acronym differently based on the Geopolitical location such that ARS gets expanded for Agricultural Revenue Service when it is from USA, but not when it is from China. Secondly, the glossaries we used for concept extraction and the thesaurus we constructed during the training process of normalization are being corrected manually at regular intervals as new entries are being automatically added through the use of NEMO by Lnx Research and others. This is labor-intensive and causes errors if there is a delay in correction. In the future, we plan to run NEMO on a much larger proportion (currently we trained on about 100K affiliation strings), correct the glossaries and stop expanding them.

**Potential Impact:** Identifying the organization of articles improves author disambiguation as it is an important indicator to know if two authors with the same name are the same person. Author disambiguation has been used for diverse applications such as building social networks, normalizing gene names and analyzing collaborations. Farkas (35) was successful in using authors' information for improving the accuracy of a baseline gene normalization system from 80% to 97%. Large scale social network analysis of disambiguated author information is useful for finding key scientific leaders who are “low publishers” in scientific journals (36). The extracted and normalized organization names can be used for similar applications. A side application for our Normalization process is in relation to the Seek Affiliation program implemented as part of the Medical Articles Record System (MARS) by NLM (23) which tries to solve the sub-problem of normalizing the affiliation string using string matching to correct errors made by OCR. This program achieved a precision of 86% and a recall of 88% on its test set of 519 articles because the affiliation strings are printed in small font and often includes superscripts with even smaller font. Since NEMO accurately extracts different fields of affiliation strings and provides a rich synonym set for the extracted organization, it could potentially supplement the Seek Affiliation program through better matching. NEMO could also be used to automatically index the PubMed citations with the normalized organization name and country for improving information retrieval.

# 5. Conclusion

In this study, we presented NEMO (Normalization Engine for Matching Organizations) for extracting organization names from bibliographic databases such as PubMed. We constructed a process to automatically build a database of normalized organization names using affiliation strings in PubMed abstracts. This could be useful for the analysis of organization social networks and other large scale analyses of scientific organization. The graphical user interface and java client library of NEMO is available at http://lnxnemo.sourceforge.net.

**List of Abbreviations**

AI Artificial Intelligence
BBN Bolt Beranek and Newman
ESS Extended Smith-Waterman Score
GPE Geo Political Entity
MARS Medical Articles Record System
MeSH Medical Subject Headings
NER Named Entity Recognition
NEMO Normalization Engine for Matching Organizations (Name of the proposed system)
NLM National Library of Medicine

| | |
|---|---|
| NSW | Non Standard Word |
| NW | Needleman-Wunsch |
| SW | Smith-Waterman |
| TSS | Tight String Similarity |
| URL | Uniform Resource Locator |

**Competing Interests**

The authors declare that they have no competing interests.

**Author(s) Contributions**

Siddhartha Jonnalagadda and Philip Topham designed NEMO. Siddhartha Jonnalagadda implemented NEMO in Java and wrote the manuscript.

**Acknowledgements**

We sincerely thank Senthil Purushothaman, Ryan Peeler, Prachi Aghi, Surendar Swaminathan, Divya Shakti, Gus Cho, Shweta Dwivedi, and Debbie Gordon (Lnx Research) who participated in the evaluation of NEMO and Sogol Amjadi (Lnx Research) who diligently copy-edited the manuscript. Thanks also to Drs. Graciela Gonzalez (Arizona State University), Dina Demner-Fushman and George Thoma (National Library of Medicine), and João Cordeiro (University of Beira Interior, Portugal) for their valuable suggestions while building NEMO. This work was entirely funded intramurally by Lnx Research.

## References

1. Treeratpituk P, Giles CL: Disambiguating authors in academic publications using random forests. In: International Conference on Digital Libraries. Austin, TX, USA: ACM; 2009 p. 39-48.
2. Torvik VI, Weeber M, Swanson DR, Smalheiser NR: A probabilistic similarity metric for Medline records: a model for author name disambiguation. In: J. Am. Information Sci. Technol. 2005; 56(2): p. 140-158.
3. Abramo G, D'Angelo CA, Pugini F: The measurement of Italian universities' research productivity by a non parametric-bibliometric methodology. Scientometrics 2008;76(2):225–244.
4. Yu W, Yesupriya A, Wulf A, Qu J, Gwinn M, Khoury MJ: An automatic method to generate domain-specific investigator networks using PubMed abstracts. BMC medical informatics and decision making 2007 Jun;7:17.
5. NextBio [http://www.nextbio.com]
6. Schmidt M, Diwersy M: Methods and Systems for Social Networking. U.S. Patent 20100017431, 2010.

7. Chinchor N, Robinson P: MUC-7 named entity task definition. In: Proceedings of the 7th Message Understanding Conference; 1997.
8. Khalid MA, Jijkoun V, Rijke MD: The impact of named entity normalization on information retrieval for question answering. Lecture Notes in Computer Science 2008;4956:705.
9. Hakenberg J, Plake C, Leaman R, Schroeder M, Gonzalez G: Inter-species normalization of gene mentions with GNAT. Bioinformatics 2008;24(16):i126.
10. Jin Y, McDonald RT, Lerman K, Mandel MA, Carroll S, Liberman MY, Pereira FC, Winters RS, White PS: Automated recognition of malignancy mentions in biomedical literature. BMC Bioinformatics 2006 Nov;7:492.
11. Annotation Guidelines for Answer Types. [http://www.ldc.upenn.edu/Catalog/docs/LDC2005T33/BBN-Types-Subtypes.html]
12. Sekine's extended named entity hierarchy. [http://nlp.cs.nyu.edu/ene/]
13. French J, Powell A, Schulman E: “Using clustering strategies for creating authority files,” Journal of the American Society for Information Science, vol. 51, 2000, pp. 774–786.
14. French J, Powell A, Schulman E, Pfaltz J: “Automating the construction of authority files in digital libraries: a case study,” Research and Advanced Technology for Digital Libraries, 1997, pp. 55–71.
15. Ganti V, Ramakrishnan R, Gehrke J, Powell A, French J: “Clustering large datasets in arbitrary metric spaces,” 15th International Conference on Data Engineering, 1999. Proceedings., 1999, pp. 502–511.
16. Galvez C, Moya-Anegón F: “The unification of institutional addresses applying parametrized finite-state graphs (P-FSG),” Scientometrics, vol. 69, 2006, pp. 323–345.
17. Jonnalagadda S, Topham P, Gonzalez G: ONER: Tool for Organization Named Entity Recognition from Affiliation Strings in PubMed Abstracts. The 3rd International Symposium on Languages in Biology and Medicine, Jeju Island, South Korea, November 8-10, 2009
18. Jonnalagadda S, Topham P, Gonzalez G: Towards Automatic Extraction of Social Networks of Organizations in PubMed Abstracts. First International Workshop on Graph Techniques for Biomedical Networks in Conjunction with IEEE International Conference on Bioinformatics and Biomedicine, Washington D.C., USA, Nov. 1-4, 2009
19. Geoworldmap database containing cities of the world with geographical coordinates [http://www.geobytes.com/freeservices.htm]
20. Fox C. A stop list for general text. ACM; 1989 p. 21.
21. Java client library for Google Translate API [http://code.google.com/p/gtranslate-java-client]
22. Sproat R, Black AW, Chen S, Kumar S, Ostendorf M, Richards C: Normalization of non-standard words. Computer speech & language(Print) 2001;15(3):287-333.
23. Hauser S, Schlaifer J, Sabir T, Demner-Fushman D, Thoma G: Correcting OCR text by association with historical datasets. In: Proc. SPIE Conference on Document Recognition and Retrieval X. 2003 p. 84-93.
24. Morgan A, Lu Z, Wang X, Cohen A, Fluck J, Ruch P, Divoli A, Fundel K, Leaman R, Hakenberg J: Overview of BioCreative II gene normalization. Genome biology 2008;9(Suppl 2):S3.
25. Barzilay R, Lee L: Learning to paraphrase: An unsupervised approach using multiple-

sequence alignment. In: proceedings of HLT-NAACL. 2003 p. 201-231.
26. Cordeiro J, Covilha P, Dias G, Cleuziou G, Orleans F: Biology Based Alignments of Paraphrases for Sentence Compression. ACL 2007 2007;:177.
27. Needleman SB, Wunsch CD: A general method applicable to the search for similarities in the amino acid sequence of two proteins. J.Mol.Biol 1970;48(3):443-453.
28. Smith TF, Waterman MS: Identification of common molecular subsequences. J.Mol.Bwl 1981;147:195-197.
29. Sequence Alignment Algorithms [http://neobio.sourceforge.net/SequenceAlignment.pdf]
30. McClosky D, Charniak E, Johnson M: Reranking and self-training for parser adaptation. In: Proceedings of the 21st International Conference on Computational Linguistics and the 44th annual meeting of the Association for Computational Linguistics. Association for Computational Linguistics Morristown, NJ, USA; 2006 p. 337-344.
31. Friedman JH: Intelligent local learning for prediction in high dimensions. In: International Conference on Artificial Neural Networks, Paris, France. 1995.
32. Levenshtein VI: Binary codes capable of correcting deletions, insertions and reversals. In: Soviet Physics Doklady. 1966 p. 707-710.
33. Cormen TH, Leiserson CE, Rivest RL, Stein C: Introduction to algorithms. New York: MIT press; 2001.
34. Manning CD, Raghavan P, Schütze H: An Introduction to Information Retrieval. Cambridge, UK: Cambridge University Press; 2008.
35. Farkas R: The strength of co-authorship in gene name disambiguation. BMC Bioinformatics 2008;9(1):69.
36. Quantity Does Not Equal Quality in Evaluating a Scientist’s Real Importance as a Key Opinion Leader [http://www.newsguide.us/technology/biotechnology/In-Key-Opinion-Leader-Management-Quantity-Does-Not-Equal-Quality-in-Evaluating-a-Scientist-s-Real-Importance]

## Supplementary File 1: Keywords for addresses

Attachment 1: Keywords_for_Addresses.xls